\documentclass[10pt,twocolumn,letterpaper]{article}

\usepackage[pagenumbers]{wacv} 

\usepackage{multirow}
\usepackage{float}
\usepackage{svg}

\definecolor{wacvblue}{rgb}{0.21,0.49,0.74}
\usepackage[pagebackref,breaklinks,colorlinks,allcolors=wacvblue]{hyperref}

\def\wacvPaperID{35} 
\def\confName{Conference}
\def\confYear{202X}

\title{Vegetation Health Classification for Low-Data and Edge Hardware Deployment}

\author{Yifei Cheng\\
Canton High School\\
Canton Township, MI\\
{\tt\small yifeichengmi@gmail.com}
\and
Haoyu Li\\
Grand Valley State University\\
Allendale, MI\\
{\tt\small lihao@gvsu.edu}
}

\begin{document}
\maketitle
\begin{abstract}
Unmanned aerial vehicles (UAVs) equipped with cameras are popular, viable tools for vegetation and environmental health monitoring. However, existing image classification methods are incompatible with UAV implementation for independent practitioners in remote, resource-constrained environments due to large dataset requirements, GPU hardware, inflexible retraining, and low interpretability. This paper presents a lightweight, interpretable vegetation classification algorithm operating with as few as five reference images per class, without GPUs or heavy machine learning frameworks. It uses a 2D hue-saturation histogram feature representation and Earth Mover's Distance or Quadratic Form Distance, paired with a log-likelihood probability score, to classify images into three vegetation health categories. We evaluated six systems: three custom pipelines and three established models across low-data scenarios ($n = 5, 10, 25$) and a full-data regime of 2,172 drone images from 28 South American locations. Our custom pipelines achieve competitive Top-1 accuracy (81–85\%) under low-data constraints, maintaining sub-1ms inference times that enable real-time classification and subsequently tasks such as seeding and watering. The algorithms also support incremental retraining without full model reconstruction. We validate edge feasibility by deploying the QFD variant on an ESP32 microcontroller, achieving 82.8\% accuracy with 1.01\% heap memory utilization ($\sim$45kb), demonstrating that effective vegetation classification is achievable on low-cost, low-power embedded hardware without cellular connectivity.
\end{abstract}
    
\section{Introduction}
\label{sec:intro}

Unmanned aerial vehicles (UAVs) with cameras and video recorders have emerged as a powerful tool for environmental monitoring due to their cost effectiveness and versatility. The role of drone imagery as a vegetation health assessment tool stretches across numerous domains and applications, such as wildfire risk monitoring, agriculture, and biodiversity conservation, especially when tools such as infrared imaging aren't available \cite{manfreda2018use, anderson2013lightweight, gonzalez2016unmanned, wang2026uav, ghali2022deep}.

Compared to the Normalized Difference Vegetation Index (NDVI), which is based on infrared imaging, pure color detection systems are less accurate in terms of measuring plant health since it actually measures physiological changes in chlorophyll density and plant stress, which are only visible in more severe stages in pure color. This makes NDVI a great choice for large scale environmental monitoring. However, in many practical scenarios independent practitioners or small organizations working on remote conservation efforts do not have access to infrared imaging, necessitating pure color classification \cite{rouse1974monitoring, pettorelli2005using}.

This is especially true for commercial drones, which are increasingly being equipped with onboard cameras and used for exactly this kind of monitoring work. These platforms are often operated in remote areas without cellular or network coverage, making it impractical or impossible to stream imagery back to a remote computer for processing. They also typically lack the dedicated GPU hardware found on higher-end research-funded platforms, and may rely on legacy microcontrollers or other low-power embedded hardware for onboard computation.

In addition, many tasks done by drones reqiure real-time classification to make decisions under field conditions, such as reforestation seeding and watering \cite{stamatopoulos2024uav, ahansal2022towards}. However, far fewer of these models simultaneously satisfy the computational and logistical constraints faced by resource-limited practitioners operating in remote field environments \cite{albanese2022low}.

To address the aforementioned practical issues with previous approaches, this paper presents a lightweight, interpretable vegetation classification algorithm designed to operate with as few as five reference images per class, without requiring GPU hardware, extensive model retraining, or heavy external machine learning frameworks. The algorithm is designed with the out-of-the-box practitioner in mind: rather than assuming access to hyper-optimized pipelines or high-performance infrastructure, it targets the hardware and workflows that independent field operators are most likely to actually have available.

Because our algorithm runs its inference entirely on-device, it is well suited to deployment directly onboard commercial drones and low-power embedded hardware, including in areas where cellular or other network signals cannot reach, removing the need to transmit imagery back to a remote computer for processing and also increasing operational range. To validate this directly, we deploy and benchmark the algorithm on an ESP32 microcontroller as a representative example of the low-cost, low-power hardware a practitioner might have on hand that has been proven to possess capabilities for computer vision processing \cite{esp32datasheet, chang2025design}.
\section{Related Work}
\label{sec:relatedWork}

This section provides an overview of existing literature, highlighting capabilities of existing small classification models while also noting constraints in these methods.

\subsection{Aerial Vegetation Monitoring and Color Spaces}

The field of real-time automated vegetation classification has grown alongside both drone and machine learning technology. In particular, supervised approaches such as Support Vector Machines (SVMs) and CNNs have demonstrated great, deployable accuracy scores for identifying vegetation health \cite{shi2015support, shi2025evaluation, jintasuttisak2025accurate}. Neural networks have also gained massive popularity recently. With the help of ImageNet, architectures such as Ultralytics' YOLOv8 are adaptable enough to fit into this domain, achieving extremely high accuracy with massive datasets \cite{jocher2023ultralytics}.

HSV was selected as the color representation method over RGB because prior research has consistently demonstrated that luminance functions as a confounding factor in RGB-based image classification \cite{song2023new, hernandez2016optimal}. Furthermore, HSV facilitates more straightforward distance calculations through eliminating the need for normalization and enabling efficient value isolation. This methodological choice is well-established within the literature \cite{sural2002segmentation, yang2015greenness, wenbo2025identification, hassanein2018new, swain1991color}.

\subsection{Deep Learning and Few-Shot Baselines in Data-Scarce Regimes}

Additionally, Few Shot Learning (FSL) methods have been proposed as a solution to low-data situations. By learning generalizable feature embeddings from related tasks or by employing metric-based similarity frameworks, FSL methods have been proven to be robust, accurately classifying vegetation scenes with as few as five labeled examples per class \cite{zhang2024convolutional, snell2017prototypical, uzhinskiy2021one}.

While these advancements represent significant steps forward, they present meaningful practical limitations that motivate the approach taken in this study.

\begin{itemize}
\item Massive data requirements: The performance of supervised deep learning and Support Vector Machines comes at the cost of a massive data requirement. They often require hundreds to thousands of annotated images to be robust, requiring many hours by experts to sort through and categorize them for training.

\item High computational expenses during training time, resultant inflexible training cycles: SVMs and neural networks require a complete retraining cycle to be completed when even minor amounts of new data is added \cite{losing2018incremental}. In field conditions where an operator might want to continuously add newly collected data to a model, this restraint is incredibly difficult to work around with.

\item Black-box neural networks: Neural networks and FSL methods essentially act as black boxes \cite{lipton2018mythos}. After training, the embeddings and internal representations of images are high-dimensional and unintuitive, and as a result essentially inaccessible to any independent practitioner trying to apply these technologies.

\item Low interpretability in SVMs: SVMs seem like a solution to this problem at first; they are more transparent and offer a more intuitive and well-understood geometric representation of images, but in most cases that involve vegetation classification, feature extraction must be programmed specifically for this domain, and a complex kernel has to be applied. For the vast majority of users applying SVMs, they essentially become a black box after applying a kernel trick due to the program now introducing infinite dimension space and no directly interpretable feature weights \cite{tharwat2019parameter}.
\end{itemize}

\subsection{Edge Deployment Constraints}

Although FSL methods largely solve the data issue, they almost always rely on deep feature extraction based off of frameworks such as ResNet and EfficientNet \cite{he2016deep, tan2019efficientnet}. While highly effective and easy to deploy, they are computationally expensive both during training and on inference time, especially when deployed on devices without dedicated GPUs, older field laptops, the processors on commercial drones, and legacy microcontrollers. For real-time assessment and analysis, this computational overhead is a serious challenge and delays conservation efforts.
\section{Methodology}
\label{sec:methodology}

This section outlines how we organized the comparative study to test our lightweight algorithm against traditional and deep learning / neural baselines. 

\subsection{Dataset Sourcing}
The dataset consists of drone-captured vegetation imagery sorted into three classes: Dry, mixed, and dense. Dry vegetation consists of essentially no healthy vegetation and is usually either completely bare or dried out. Mixed vegetation images have some mixture of healthy green vegetation together with dried out vegetation or bare dirt. Dense vegetation images are dominated by green, lush vegetation.

\begin{figure}[H]
    \centering
    \includegraphics[width=\columnwidth]{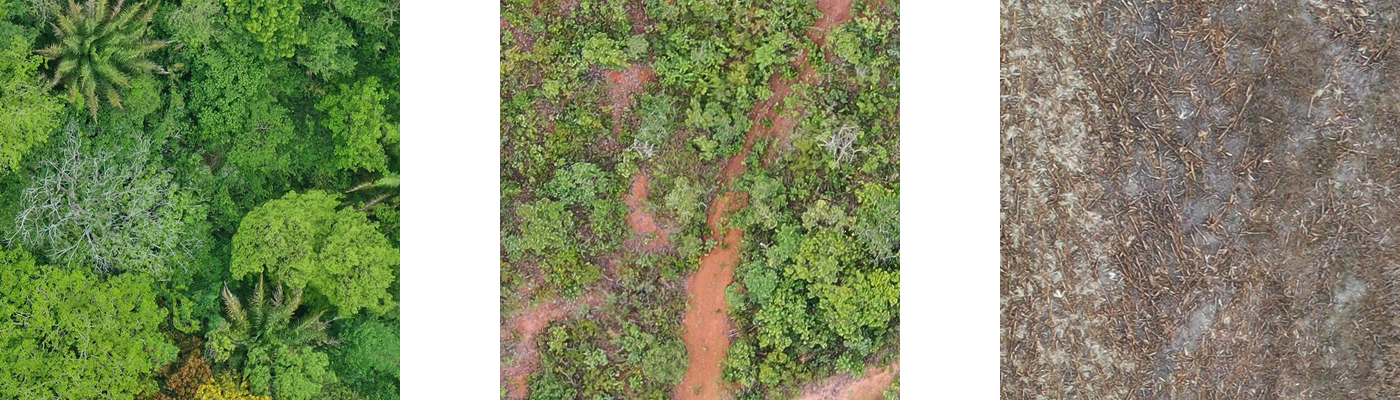}
    \caption{Archetypical examples of the three vegetation classes 
    used in this study. Left: Dense. Center: Mixed. Right: Dry.}
    \label{fig:vegetation_samples}
\end{figure}

There are a total of 2172 images in the dataset. 783 of them are dense vegetation, 625 are dry, and 764 are mixed. The drone imagery is taken from 28 different locations within South America from OpenAerialMap \cite{openaerialmap}. 21 of these locations were sorted for all three classes, and an additional 7 locations were sorted specifically to add to the dry class due to a severe dataset size mismatch in comparison to the mixed and dense classes. Out of the dry images, 330 come from the original 21 locations, and the additional 295 were supplemented from the additional locations. 

All 2172 images were manually classified by the first author to establish ground truth. Ambiguous, low-quality, or nondescript images were excluded.

\subsection{Feature Representation and Extraction}

All systems except for the Siamese Network, utilize the same standardized feature representation layer. This shared foundation isolates algorithm-level differences in performance and accuracy from variation based on feature extraction.

The 2D histogram employed by four of the five systems evaluated comprises 15 hue columns and 8 saturation rows. Each hue column covers 24 degrees of the color wheel, while each saturation row covers 12.5\% of the saturation spectrum.

When constructing the histogram, pixels with values exceeding 240 or falling below 20 are excluded from consideration. These extreme values are perceptually ambiguous and usually just come from camera glare, deep shadows, or image processing artifacts. As a result, they contain very little real information about the hue composition of an image and are disregarded.

Finally, each pixel is assigned to the corresponding histogram cell, and the resulting histogram is normalized to account for variations introduced by the value threshold and potential differences in image dimensions. This feature histogram is shared across all systems.

Prior to inference, both EMD algorithms prune the histogram for values under 0.5\% and renormalize the histogram in order to avoid slowing down the algorithm with more bins with tiny values that largely do not end up influencing the final perceptual composition of the image.

The Siamese Network purely utilizes a CNN as a feature extraction method since it is the industry standard feature extraction pairing for Siamese networks (Although technologies such as Vision Transformers are emerging as more popular, state-of-the-art alternative for CNNs that require more data and compute power) and CNNs were specifically made for the purpose of image analyzation.

\subsection{Similarity Metrics and Mathematical Framework}

Earth Movers Distance (Abbreviated as EMD) and Quadratic Form Distance (Abbreviated as QFD) are the two cross-bin distance metrics of choice for the custom algorithm, both respecting the shape of the histogram distributions rather than calculating distance based on isolated, bin-by-bin comparisons \cite{hafner1995efficient, rubner2000earth}.

$$
EMD(P, Q) = \frac{\min_{F \in \mathcal{F}} \sum_{i} \sum_{j} f_{ij} d_{ij}}{\sum_{i} \sum_{j} f_{ij}}
$$
,where $F = [-f_{ij}]$ is the optimal flow between bins of histograms $P$ and $Q$, and $d_{ij}$ is the perceptually weighted ground distance.

EMD treats each histogram as a distribution of "mass" spread across bins and computes the minimum cost required to reshape or move one distribution into the other via a linear programming problem, calculating cross-bin perceptual distance where cost is a function of how far and how much mass travels \cite{ling2007efficient}.

However, although EMD is incredibly accurate at capturing cross-bin distribution similarity, the linear programming approach scales cubicly and requires a high memory footprint since the optimization needs to search over many possible reassignments of mass, which becomes expensive as histogram resolution increases.

$$
QFD(P, Q) = \sqrt{(P - Q)^T A (P - Q)}
$$
,where $A$ is a similarity matrix with entries $a_{ij} = 1 - (d_{ij} / d_{\max})$.

QFD approximates this same cross-bin sensitivity without an iterative solver. Instead of optimizing a flow, it directly computes a weighted distance between the two histogram vectors, where the weighting matrix encodes how similar any given pair of bins is expected to be. Because this reduces to a fixed matrix multiplication rather than a search process, QFD is far cheaper to compute and far more predictable in memory usage, at the cost of using a fixed, precomputed notion of bin-to-bin similarity rather than solving for the true optimal transport each time. Although EMD is more robust at handling variable distribution sizes and dramatic feature shifts, QFD is an efficient alternative for ESP32 testing.

Within the distance calculation, a polar mapping ensures that hue differences in lower saturation bins are penalized less than in highly saturated bins, reflecting the perceptual ambiguity of low-saturation colors. For histogram bin $i$ with hue index $h_i$ and saturation index $s_i$ (out of $N_H$ hue bins and $N_S$ saturation bins), we first map the bin into a 2D polar coordinate. The ground distance $d_{ij}$ between bins $i$ and $j$ used in both EMD and QFD is then the Euclidean distance between their polar-mapped coordinates:

$$
d_{ij} = \sqrt{(x_i - x_j)^2 + (y_i - y_j)^2}
$$

Since $r_i$ scales linearly with saturation, bins with low saturation map close to the origin regardless of their hue angle $\theta_i$, compressing the effective distance between hues at low saturation. This distance is finally normalized by the maximum pairwise distance $d_{\max}$ observed across all bin pairs, giving the similarity weighting $a_{ij} = 1 - (d_{ij}/d_{\max})$ used in the QFD similarity matrix $A$.

\begin{figure}[H]
    \centering
    \includegraphics[width=0.75\linewidth]{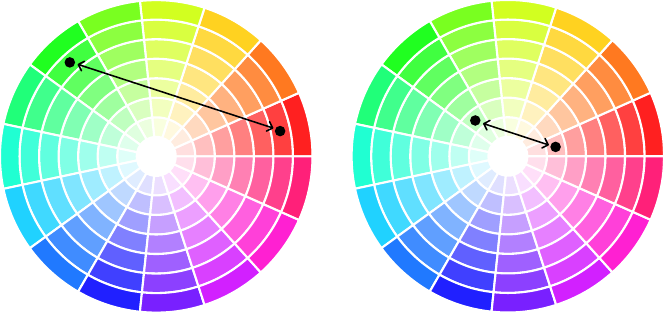}
    \caption{Visualization of radial saturation calculation. Bins with identical relative hue placements are penalized less with lower saturation due to being closer to the origin, reflecting real-life perceptual ambiguity}
    \label{fig:placeholder}
\end{figure}

Alongside the geometric distance metrics, the pipeline design incorporates a log-likelihood term to help disambiguate images that may sit at similar geometric distances from class archetype histograms but differ in how their probability mass is distributed across bins.

The averaged reference histogram for each class is treated as an empirical probability distribution over hue-saturation bins. Taking its natural log (with a small epsilon to prevent log(0)) converts it into per-bin log-probabilities. The test histogram is then scored against each class via a dot product.

Since both histograms are normalized, this is mathematically equivalent to negative cross-entropy, which is asking how likely the test image's color distribution would be if drawn from each class. This score is min-max normalized and combined with the distance score, making the two terms act as independent votes. The distance metric captures distributional shape difference, while the log-likelihood captures class fit. During ablation testing, combining both was found to stabilize predictions against the mixed class, whose intermediate color profile can otherwise pull single-metric classifiers toward it, raising Top 1 Accuracy by around $5\%$.

\section{Implementation \& Evaluation}
\label{sec:implementationAndEvaluation}

This section details the implementation of each system as used in this research and how each will be evaluated. To keep the comparative analysis fair, all six systems are tested using the same image data across three low data constraints (n=5, n=10, and n=25 reference images per class) and one full data situation. These constraints help simulate early conditions faced by independent practitioners when they might not have access to extensive training data. Additionally, to evaluate real-world feasibility on edge hardware, all systems are tested across two test sessions: one on Google Colab and one on an ESP32 microcontroller.

\subsection{Tested Systems}

We evaluate six system configurations using identical sample sizes: three custom histogram distance pipelines and three standard machine learning baselines. To reflect field conditions for environmental practitioners without high-performance computing infrastructure, baselines utilize standard out-of-the-box configurations. Additionally, all systems perform unbatched, single-image inference to simulate real-time on-the-fly streaming over low-bandwidth connections.

\begin{itemize}
    \item \textbf{EMD-Average (Custom):} Computes an average 2D reference histogram across $n$ sample images per class, classifying test images using Earth Mover's Distance (EMD).
    
    \item \textbf{EMD-Median (Custom):} Identical pipeline to EMD-Average, but uses a bin-wise median reference histogram to provide outlier resistance against class distribution skew.

    \item \textbf{QFD Variant (Custom):} Shares the shared 2D histogram representation but replaces EMD with Quadratic Form Distance (QFD) for hardware efficiency.

    \item \textbf{Siamese Network Baseline:} A few-shot neural network using a four-block CNN encoder backbone (32, 64, and 128 channels) projecting into a 64-dimensional $L_2$-normalized embedding space \cite{bromley1993signature}. Trained with contrastive loss (margin 1.0, Adam $lr=10^{-3}$, batch size 32, 30 epochs) on balanced 50/50 positive and negative pairs. At inference, it acts as a prototypical classifier, assigning test image embeddings to the nearest class prototype calculated from average reference embeddings.

    \item \textbf{Random Forest Baseline:} A \texttt{scikit-learn} forest of 200 estimators with balanced class weighting trained directly on the flattened 120-bin histogram \cite{breiman2001random}. Texture channels (LBP) were omitted as preliminary testing showed marginal accuracy improvements (+0.71\%) at prohibitive latency costs.

    \item \textbf{SVM Baseline:} A standard \texttt{scikit-learn} SVM with an RBF kernel and standard feature scaling, trained on the shared 120-element flattened histogram with balanced class weighting \cite{suthaharan2016support}.
\end{itemize}

\subsection{Preliminary Development \& Design Rationale}

Iterative development guided the algorithm's architecture and feature selection. Initial tests using a coarse $8 \times 4$ hue-saturation histogram suffered from high classification bias toward the intermediate mixed class. Increasing resolution to $15 \times 8$ resolved this bias by establishing tighter class boundaries.

While EMD-Average and EMD-Median leverage cross-bin properties, preliminary hardware profiling revealed critical edge deployment barriers. EMD's iterative linear programming and lack of vectorizability create severe memory overhead and sequential computation bottlenecks on the ESP32 microcontroller, compounded by the hardware's lack of a floating-point divider. This necessitated the QFD Variant, which computes all distances simultaneously via matrix operations.

In evaluating feature sets, adding Local Binary Pattern (LBP) texture channels (sampled at radii 1, 3, and 5) to the Random Forest and SVM yielded a negligible top-1 accuracy gain (+0.71\%) while increasing end-to-end latency by 100~ms, prompting its exclusion. A naive RGB baseline with $L_2$ distance achieved only 50.84\% top-1 accuracy at $n=5$ and plateaued at 39\% mixed-class accuracy at $n=25$ ($\sigma=8.1\%$), confirming the inadequacy of basic color averaging.

To resolve ambiguous geometric distances among classes with distinct probability distributions, a log-likelihood scoring term was integrated into the custom pipelines, yielding a 5\% top-1 accuracy improvement in ablation testing with minimal latency. Finally, profiling revealed that sparse, near-zero histogram bins severely bottlenecked OpenCV's EMD solver ($O(K^3 \log K)$ complexity). Implementing a 0.5\% bin pruning threshold reduced per-image inference time from 16~ms to sub-1~ms without negatively altering baseline metrics.

\subsection{Experimental Procedure and Edge Validation}
To simulate real-world field or data-scarce conditions, we extract a permanent, isolated testing set, made up of 20\% of the images from each class. The test set was a stratified sample to ensure that the slight class size mismatch wouldn't cause any issues. The remaining images form the training pool. For each test run, we sample exactly n images per class to act as the training/reference data. Every experiment configuration is mapped and executed during our primary benchmarking platform (Google Colab), and the ESP32-optimized QFD Variant on an ESP32 microcontroller.

Because low-data training is highly sensitive to outlier images and variation, each system for each data constraint is tested 20 times with n randomly chosen images. The final metrics for these systems represent the average across all runs. For the full-data scenarios, each system was only tested on one round since additional rounds would contain the same training data and would therefore be redundant.

\subsubsection{Cloud Benchmarking Platform (Google Colab)}
All training and inference testing in Google Colab is done on the CPU for fairness and to match domain needs.

\subsubsection{Edge Device Deployment (ESP32 Microcontroller)}
To validate that our algorithm runs on low-end hardware without GPU, we test directly on an ESP32 microcontroller, which will serve as a representative platform for legacy, low-power hardware found on commercial drone flight controllers. Preliminary evaluation showed EMD's iterative linear programming is incompatible with ESP32 resources, so edge validation will test the QFD Variant, which shares the feature extraction pipeline but uses predictable matrix operations that conform to the ESP32's hardware limitations (See section Preliminary Development).

Only the QFD Variant is tested on ESP32 at n=5 to demonstrate feasibility. Since QFD stores exactly three reference histograms regardless of n (one per class), memory usage is invariant to training sample count, making higher-n testing redundant. Accuracy scaling is already characterized by Google Colab results.

Training occurs in Google Colab; only inference is evaluated on ESP32. Reference histograms are compiled and flashed onto the device to simulate real-world deployment. The 15 training images (3 groups of 5) were randomly chosen for one trial.

\subsection{Evaluation Metrics}
Each system is benchmarked against nine core performance metrics on our main hardware platform. Of these, only the metrics related to inference (Total Accuracy, Partial Accuracy, Inference Time, Compute Power during Inference Time, and RAM Usage during Inference Time) are also measured during the ESP32 session.
\begin{itemize}
    \item Total Accuracy (Top-1): The percentage of test images placed into exactly the correct class.
    \item Partial Accuracy: The percentage of test images placed into either the exact right class or an adjacent class. Because our classes are linear (Dry, Mixed, Dense), misclassifying a dense image as mixed is treated as a partial success, whereas misclassifying dense as dry is a total failure.
    \item Inference Time per Image: The average time (in milliseconds) required to process and classify a single input image after the model is established.
    \item Training time per round (Only tested on main platform)
    \item Memory usage during inference (Only tested on ESP32)
\end{itemize}
Feature extraction time and data ingestion time is also isolated and measured in Google Colab. Since all except for the Siamese Network utilize the same feature extraction method and ingests data at the same rate, they will not be taken into consideration during comparative benchmarking, and will be noted in results independently.
\section{Results}

\subsection{Accuracy Under Low Data Constraints}

Across low-data regimes, the custom pipelines (EMD-Average, EMD-Median, and QFD Variant) consistently achieve competitive or superior Top-1 accuracy over neural and kernel baselines, demonstrating small sample efficiency at $n = 5$ and $n = 10$. 

This low-data advantage narrows as $n$ increases, with Random Forest eventually surpassing all custom variants at $n = 25$. EMD-Average and the QFD Variant exhibit nearly identical performance at the smallest sample size before plateauing near ~85\% Top-1 accuracy. Something interesting is that EMD-Average consistently suffers from severe class imbalance, with dry and dense classes having significantly (~25\%) more accurate placements compared to mixed.

The SVM struggles most severely under extreme data scarcity at $n = 5$, confirming that kernel-based classifiers require broader training support to generalize, while its low partial accuracy highlights a susceptibility to catastrophic misclassification. The Siamese Network displays similar low-$n$ fragility before recovering strongly at $n = 25$. Random Forest proves to be the most robust baseline, overtaking the custom pipelines at higher sample counts. In addition, partial accuracy remains uniformly high across all six systems and data regimes. They reach generally 95\% and reaching 100\% for custom pipelines at higher $n$, which hows that high-$n$ misclassifications reflect benign boundary shifts into neighboring classes rather than severe model failures.

\begin{table}[H]
\centering
\footnotesize
\begin{tabular}{llll}
\toprule
                                 & n = 5                       & n = 10                      & n = 25                       \\
\midrule
\multirow{2}{*}{EMD Averaging}   & 81.1\% ± 4.6\%& \textbf{85.0\% ± 2.8\%}& \textbf{85.1\% ± 1.8\%}\\
                                 & \multicolumn{1}{r}{\textbf{100\%}} & \multicolumn{1}{r}{\textbf{100\%}}   & \multicolumn{1}{r}{\textbf{100\%}}    \\
\midrule
\multirow{2}{*}{EMD Median}      & 74.1\% ± 6.0\%& 80.9\% ± 2.9\%& 81.4\% ± 2.6\%\\
                                 & \multicolumn{1}{r}{\textbf{100\%}} & \multicolumn{1}{r}{\textbf{100\%}} & \multicolumn{1}{r}{\textbf{100\%}}  \\
\midrule
\multirow{2}{*}{QFD Variant}     & \textbf{81.4\% ± 4.6\%}& 82.7\% ± 2.3\%& 83.4\% ± 1.4\%\\
                                 & \multicolumn{1}{r}{99.97\%} & \multicolumn{1}{r}{\textbf{100\%}}   & \multicolumn{1}{r}{\textbf{100\%}} \\
\midrule
\multirow{2}{*}{Siamese Network} & 67.3\% ± 9.7\%& 76.3\% ± 5.8\%& 84.3\% ± 5.5\%\\
                                 & \multicolumn{1}{r}{98.56\%} & \multicolumn{1}{r}{99.69\%} & \multicolumn{1}{r}{99.96\%}  \\
\midrule
\multirow{2}{*}{Random Forest}   & 79.3\% ± 5.2\%& 85.3\% ± 4.6\%& 89.0\% ± 2.3\%\\
                                 & \multicolumn{1}{r}{94.00\%} & \multicolumn{1}{r}{98.00\%} & \multicolumn{1}{r}{\textbf{100\%}}     \\
\midrule
\multirow{2}{*}{SVM}             & 61.8\% ± 6.1\%& 70.5\% ± 3.8\%& 77.3\% ± 2.4\%\\
                                 & \multicolumn{1}{r}{81.00\%} & \multicolumn{1}{r}{82.00\%} & \multicolumn{1}{r}{90\%}  
                                \\ \bottomrule
\end{tabular}
  \caption{Top 1 macro accuracy ± standard deviation (Above) and partial accuracy (Below)}
  \label{tab:lowDataAccuracy}
\end{table}

\subsection{Inference and Training Time Under Low Data Constraints}

With the exception of the Siamese Network baseline, all systems utilize the identical 2D Hue-Saturation histogram feature extractor, resulting in a consistent feature extraction time per image of \~26 ± 7 ms. For the EMD and QFD pipelines, this feature extraction time is the primary bottleneck. The Siamese Network utilizes a traditional CNN feature extractor, among the most common methods of feature extraction paired with this baseline.

For $n = 5, 10, and 25$, all three custom pipelines achieve the shortest inference and training time in comparison to the three baselines. 

For the Siamese network, the flat inference time across n values is a direct consequence of how the CNN operates: at inference, the network performs a single forward pass through a fixed convolutional architecture to produce an embedding vector, and then computes a similarity score against all of the reference embeddings. Since the forward pass cost depends only on image resolution and network depth instead of how many reference images were used during training, adding more reference samples does not increase per-image inference latency. This contrasts with the histogram-based systems, where inference involves computing EMD or QFD distances against aggregated reference histograms, though in practice these operations remain fast due to the compact histogram representation.

\begin{table}[H]
\centering
\footnotesize
\setlength{\tabcolsep}{3pt}
\begin{tabular}{llll}
\toprule
                  & n = 5           & n = 10          & n = 25          \\
\midrule
EMD Averaging     & 0.939 ± 0.508& 1.009 ± 0.530& 1.013 ± 0.519\\
\midrule
EMD Median        & 0.741 ± 0.400& 0.769 ± 0.325& 0.734 ± 0.309\\
\midrule
QFD Variant       &\textbf{0.315 ± 0.074}   & \textbf{0.316 ± 0.084}   & \textbf{0.314 ± 0.086}   \\
\midrule
Siamese Network   & 19.896 ± 4.139  & 19.896 ± 4.139  & 20.067 ± 4.281  \\
\midrule
Random Forest     & 16.196 ± 5.132   & 15.919 ± 4.384  & 15.990 ± 4.701  \\
\midrule
SVM               & 1.089 ± 0.351   & 1.076 ± 0.234   & 1.092 ± 0.241   \\
\bottomrule
\end{tabular}
\caption{Inference time per image ± Standard deviation (ms)}
\label{tab:lowDataInferenceTimes}
\end{table}

For all three custom pipeline variants, training time scales linearly in accordance with $n$ since additional images take the same amount of time to aggregate into the histogram for training, yet remain largely manageable. The SVM keeps pace with the custom pipeline variants, and all four systems remain consistently the fastest to train across all $n$ values.

The Random Forest is the second slowest system for training, scaling from 1.110 ± 0.201 seconds at $n=5$ to 4.035 ± 0.434 seconds at $n=25$, slightly above the custom pipelines and SVM at comparable $n$ values. 

The Siamese Network is the slowest by a substantial margin, reaching 173.9 ± 4.372 seconds at just $n=5$ and rapidly scaling up as $n$ increases. This reflects the cost of fine-tuning or re-training the convolutional architecture on CPU without GPU acceleration, where backpropagation through the network layers grows proportionally with the number of training samples. The result is that even at small values of $n$, the Siamese Network's training overhead exceeds that of every other system by roughly two orders of magnitude.

\begin{table}[H]
\centering
\footnotesize
\begin{tabular}{llll}
\toprule
                  & n = 5           & n = 10          & n = 25          \\
\midrule
EMD Averaging     & 0.721 ± 0.152& 1.483 ± 0.224& 3.514 ± 0.457\\
\midrule
EMD Median        & \textbf{0.702 ± 0.121}& \textbf{1.397 ± 0.218}& 3.505 ± 0.452\\
\midrule
QFD Variant       & 0.735 ± 0.114   & 1.539 ± 0.256   & 3.667 ± 0.495   \\
\midrule
Siamese Network   & 173.9 ± 4.372 & 347.5 ± 4.012 & 892.8 ± 7.684 \\
\midrule
Random Forest     & 1.110 ± 0.201   & 1.824 ± 0.274   & 4.035 ± 0.434   \\
\midrule
SVM               & 0.763 ± 0.174   & 1.443 ± 0.197   & \textbf{3.480 ± 0.272}   \\
\bottomrule
\end{tabular}
\caption{Training time per group of $n$ ± standard deviation (sec)}
\label{tab:lowDataTrainingTimes}
\end{table}

\subsection{Full Data Environment}

In the full-data setting, the accuracies observed at low $n$ reverses: all three baselines now outperform the custom pipelines on Top-1 accuracy, with Random Forest (94.67\%) and the Siamese Network (93.25\%) achieving the highest accuracy, followed by SVM (90.84\%). The custom pipelines plateau in a tight band around 83–85\% regardless of the distance metric used, indicating that the 2D hue-saturation histogram representation itself, rather than the choice of EMD-Averaging, EMD-Median, or QFD, is the limiting factor once enough labeled data is available to fully exploit the higher-capacity baselines. 

In the full data scenario, although all systems maintain a 100\% partial accuracy, all baselines perform better in terms of Top-1 accuracy compared to our custom systems. Custom systems plateau around 85\% Top-1 accuracy. 

Notably, partial accuracy remains at 100\% for every system in the full-data regime, so none of the added baseline accuracy comes at the expense of severe misclassifications.

\begin{table}[H]
\centering
\footnotesize
\begin{tabular}{lrr}
\toprule
              & \multicolumn{1}{l}{Top1} & \multicolumn{1}{l}{Partial} \\
\midrule
EMD Averaging & 86.05\%& 100\%                       \\
EMD Median    & 82.82\%& 100\%                       \\
QFD Variant   & 85.29\%                  & 100\%                       \\
Siamese Net.  & 93.25\%                  & 100\%                       \\
Random Forest & \textbf{94.67\%}                  & 100\%                       \\
SVM           & 90.84\%                  & 100\%                       \\
\bottomrule
\end{tabular}
\caption{Model accuracy results}
\label{tab:fullDataAccuracy}
\end{table}

Training time diverges sharply: the Siamese Network requires 20014.5 seconds (over 5.5 hours), compared to under 110 seconds for every other system. QFD and both EMD systems keep the trend of maintaining the fastest inference out of hte six systems. Both EMD-Averaging and EMD-Median also achieve sub minute training times per group, the fastest between all six groups.

In terms of average inference time for all images, SVM performs the fastest out of the comparison baselines. Although both Siamese Network and Random Forest perform the slowest in terms of average inference time, even when taking into account feature extraction (\~26ms), on comparable hardware both can theoretically achieve around 20 fps in a real time practical scenario, still making them viable for deployment.

The QFD Variant maintains the fastest inference (<0.4ms) in the all-data environment, and Systems EMD Average and EMD Median incur higher inference costs due to EMD's iterative optimization, even with the pruning system. For all non-neural systems, training times remain tightly clustered (under 5 seconds at n=25), enabling practical field retraining. Although performing accurately, the 5.5 hour retraining cycle on a CPU for the Siamese Network indicates that full retraining cycles would be difficult to integrate into low-end situations for practitioners.

\begin{table}[H]
\centering
\footnotesize
\begin{tabular}{lll}
\toprule
              & Training Time (sec)& Average Inf. (ms)\\
\midrule
EMD Averaging & 80.187 ± 1.225& 1.003 ± 0.528\\
EMD Median    & 79.937 ± 0.577& 0.735 ± 0.311\\
QFD Variant   & 81.865 ± 0.567& \textbf{0.341 ± 0.261}\\
Siamese Net.  & \multicolumn{1}{l}{20014.526}& 19.984 ± 3.908  \\
Random Forest & \textbf{67.892 ± 0.620}               & 16.004 ± 4.226  \\
SVM           & 91.013 ± 1.457               & 1.280 ± 0.338   \\
\bottomrule
\end{tabular}
\caption{Training and average inference times}
\label{tab:fullDataTiming}
\end{table}

\subsection{ESP32 Edge Deployment}

To validate resource-constrained feasibility, the QFD Variant was deployed on an ESP32 at n=5, achieving 82.8\% overall accuracy while utilizing just 45,136 bytes of heap memory (1.01\% utilization) with a fragmentation rate of 6.41\%.

Per image timing benchmarked on the ESP32 confirms this efficiency, but also points out a critical real-life bottleneck for these types of applications. The vast majority of the 326.09 ms end-to-end pipeline time is taken up by serial data transfer and overhead, while inference takes only around 20 ms. 

\begin{table}[H]
\centering
\footnotesize
\begin{tabular}{ll}
\toprule
End-to-End Latency              & 326.09ms \\
\midrule
Inference                       & 23.117ms \\
\midrule
Serial data transfer + Overhead & 302.97ms \\
\bottomrule
\end{tabular}
  \caption{ESP32 Latency Breakdown}
  \label{tab:esp32Timing}
\end{table}

This breakdown demonstrates that the QFD Variant itself is not the constraint on real-time performance; inference accounts for only about 7\% of the total pipeline latency. The remaining ~93\% is attributable to the serial link between the host computer and the microcontroller, suggesting that future deployments using a faster interface (e.g., SPI or onboard camera capture rather than serial streaming) or an onboard camera module could reduce end-to-end latency substantially without any change to the underlying classification algorithm. Even with this overhead included, the full pipeline completes in roughly a third of a second per image on hardware with only 520KB of internal SRAM, reinforcing the feasibility of deploying this method on low-power, low-cost field devices.

\subsection{Retraining Efficiency}

Unlike traditional machine learning models, the proposed EMD-Average and QFD pipelines do not require full retraining when new data is acquired, only requiring reaggregating the 2D hue-saturation reference histograms by weighting the incoming images against the existing class totals. Operators can perform these incremental updates directly in the field without storing raw dataset images or re-running optimization passes, offering a distinct operational advantage under time-stressed or data-constrained environments.

In contrast, the evaluation baselines introduce significant retraining overhead:
\begin{itemize}
    \item \textbf{SVM:} RBF-kernel SVMs must evaluate pairwise interactions across all data points, requiring a complete retrain from scratch whenever new samples are introduced.
    \item \textbf{Random Forest:} While new decision trees can technically be appended, balancing decision weights between old and new trees is difficult, and modifying existing trees without full retraining is impractical for field deployment.
    \item \textbf{Siamese Network:} Adding reference samples requires fine-tuning or full end-to-end retraining to preserve the global embedding space. Given the sub-optimal CPU training latency observed in Section 4.3 (over 5.5 hours), rolling field updates are not very viable..
\end{itemize}
\section{Discussion \& Future Work}
\label{sec:discussion}

We evaluated lightweight 2D hue-saturation histogram classification pipelines against three commonly used baselines across low-data ($n \le 5$) and full-data regimes. Our goal was to assess whether cross-bin metrics paired with probability scoring are capable of real-time performance under extreme data scarcity, zero GPU hardware, incremental updates, and FPU-less microcontrollers.

Under low-data constraints ($n=5$), the proposed EMD-Averaging and QFD pipelines achieves ~81\% Top-1 accuracy matching or exceeding all baselines with sub-1ms CPU inference, enabling tasks that demand real-time classification such as forest seeding regeneration and watering for further action. Their reaggregation scheme also allows training data to be updated without retraining. Conversely, SVM struggled at low sample sizes (61.8\% at $n=5$), and the Siamese Network’s high training overhead (~900s at $n=25$ on CPU) prevents field updates.

In the full data scenario, high-capacity models (Random Forest at 94.67\%, Siamese at 93.25\%) outperformed our pipelines (83–86\%). This plateau stems from the 2D HSV representation, which trades fine discriminative resolution and lack of depth for computational efficiency, a necessary tradeoff under certain data constraints. ESP32 edge deployment validated this approach. The QFD pipeline approach achieved 82.8\% Top-1 accuracy while utilizing 45 KB of heap memory (1.01\% utilization). Latency profiling revealed that a primary real-time bottleneck stems from serial communication rather than classification computation.

To address current representation limits in our comparative analysis portion, future architectural investigations can explore Prototypical networks as an alternative few-shot baseline that unifies metric learning with direct class prototype representation. We can also pursue class expansion beyond the current three class system and explicitly mitigate the mixed class imbalance caused by transitional intermediary ground types. Future work should also focus on refining our methodology and testing by testing on separate, unseen sources to more effectively identify model robustness.

To support real-world deployment, evaluation can also expand to multi-region generalization and seasonal robustness tests across diverse vegetative conditions. In addition, future work can focus on transitioning from stand-alone ESP32 benchmarking to full integration with production drone Flight Control Systems, observing practical real-time onboard inference directly and also potentially integration with downstream field tasks, such as autonomous regenerative seeding and targeted watering.

\section{Conclusion}
\label{sec:conclusion}

This paper presented a lightweight, interpretable vegetation classification pipeline combining 2D hue-saturation histograms, polar ground-distance mapping, cross-bin metrics (EMD/QFD), and log-likelihood scoring. By achieving ~81\% Top-1 accuracy under extreme low-data regimes ($n=5$) with sub-1 ms CPU inference and a 45 KB heap footprint on ESP32 hardware, the framework can provide insight on a potential approach on bridging the critical gap between real-time performance and extreme resource constraints. In adition, the reaggregation architecture enables instant field-side updates without extensive retrain cycles, offering a retrainable alternative to heavy neural networks on FPU-less edge devices. Future work should expand the class taxonomy, evaluate cross-region seasonal robustness, and integrate the pipeline directly into drone flight control systems for closed-loop, real-time tasks such as autonomous seeding and targeted watering.
{
    \small
    \bibliographystyle{ieeenat_fullname}
    \nocite{*}
    \bibliography{main}

\begin{thebibliography}{40}
\providecommand{\natexlab}[1]{#1}
\providecommand{\url}[1]{\texttt{#1}}
\expandafter\ifx\csname urlstyle\endcsname\relax
  \providecommand{\doi}[1]{doi: #1}\else
  \providecommand{\doi}{doi: \begingroup \urlstyle{rm}\Url}\fi

\bibitem[ope()]{openaerialmap}
Openaerialmap.
\newblock \url{https://openaerialmap.org}.
\newblock Accessed: 2026.

\bibitem[Ahansal et~al.(2022)Ahansal, Bouziani, Yaagoubi, Sebari, Sebari, and Kenny]{ahansal2022towards}
Youssef Ahansal, Mourad Bouziani, Reda Yaagoubi, Imane Sebari, Karima Sebari, and Lahcen Kenny.
\newblock Towards smart irrigation: A literature review on the use of geospatial technologies and machine learning in the management of water resources in arboriculture.
\newblock \emph{Agronomy}, 12\penalty0 (2):\penalty0 297, 2022.

\bibitem[Albanese et~al.(2022)Albanese, Nardello, and Brunelli]{albanese2022low}
Andrea Albanese, Matteo Nardello, and Davide Brunelli.
\newblock Low-power deep learning edge computing platform for resource constrained lightweight compact uavs.
\newblock \emph{Sustainable Computing: Informatics and Systems}, 34:\penalty0 100725, 2022.

\bibitem[Anderson and Gaston(2013)]{anderson2013lightweight}
Karen Anderson and Kevin~J Gaston.
\newblock Lightweight unmanned aerial vehicles will revolutionize spatial ecology.
\newblock \emph{Frontiers in Ecology and the Environment}, 11\penalty0 (3):\penalty0 138--146, 2013.

\bibitem[Azmi et~al.(2025)Azmi, Gibran, and Saleh]{azmi2025herbal}
Fadhillah Azmi, M~Khalil Gibran, and Amir Saleh.
\newblock Herbal plant image retrieval using hsv color histogram and random forest algorithm.
\newblock \emph{Journal of Computer Science, Information Technology and Telecommunication Engineering}, 6\penalty0 (2):\penalty0 1019--1027, 2025.

\bibitem[Breiman(2001)]{breiman2001random}
Leo Breiman.
\newblock Random forests.
\newblock \emph{Machine learning}, 45\penalty0 (1):\penalty0 5--32, 2001.

\bibitem[Bromley et~al.(1993)Bromley, Guyon, LeCun, S{\"a}ckinger, and Shah]{bromley1993signature}
Jane Bromley, Isabelle Guyon, Yann LeCun, Eduard S{\"a}ckinger, and Roopak Shah.
\newblock Signature verification using a" siamese" time delay neural network.
\newblock \emph{Advances in neural information processing systems}, 6, 1993.

\bibitem[Chang et~al.(2025)Chang, Wu, and Lin]{chang2025design}
Yeong-Hwa Chang, Feng-Chou Wu, and Hung-Wei Lin.
\newblock Design and implementation of esp32-based edge computing for object detection.
\newblock \emph{Sensors}, 25\penalty0 (6):\penalty0 1656, 2025.

\bibitem[{Espressif Systems}(2024)]{esp32datasheet}
{Espressif Systems}.
\newblock \emph{{ESP32-WROVER-E \& ESP32-WROVER-IE Datasheet}}, 2024.
\newblock Version 1.4.

\bibitem[Ghali et~al.(2022)Ghali, Akhloufi, and Mseddi]{ghali2022deep}
Rafik Ghali, Moulay~A Akhloufi, and Wided~Souidene Mseddi.
\newblock Deep learning and transformer approaches for uav-based wildfire detection and segmentation.
\newblock \emph{Sensors}, 22\penalty0 (5):\penalty0 1977, 2022.

\bibitem[Gonzalez et~al.(2016)Gonzalez, Montes, Puig, Johnson, Mengersen, and Gaston]{gonzalez2016unmanned}
Luis~F Gonzalez, Glen~A Montes, Eduard Puig, Sandra Johnson, Kerrie Mengersen, and Kevin~J Gaston.
\newblock Unmanned aerial vehicles (uavs) and artificial intelligence revolutionizing wildlife monitoring and conservation.
\newblock \emph{Sensors}, 16\penalty0 (1):\penalty0 97, 2016.

\bibitem[Hafner et~al.(1995)Hafner, Sawhney, Equitz, Flickner, and Niblack]{hafner1995efficient}
James Hafner, Harpreet~S. Sawhney, William Equitz, Myron Flickner, and Wayne Niblack.
\newblock Efficient color histogram indexing for quadratic form distance functions.
\newblock \emph{IEEE transactions on pattern analysis and machine intelligence}, 17\penalty0 (7):\penalty0 729--736, 1995.

\bibitem[Hassanein et~al.(2018)Hassanein, Lari, and El-Sheimy]{hassanein2018new}
Mohamed Hassanein, Zahra Lari, and Naser El-Sheimy.
\newblock A new vegetation segmentation approach for cropped fields based on threshold detection from hue histograms.
\newblock \emph{Sensors}, 18\penalty0 (4):\penalty0 1253, 2018.

\bibitem[He et~al.(2016)He, Zhang, Ren, and Sun]{he2016deep}
Kaiming He, Xiangyu Zhang, Shaoqing Ren, and Jian Sun.
\newblock Deep residual learning for image recognition.
\newblock In \emph{Proceedings of the IEEE conference on computer vision and pattern recognition}, pages 770--778, 2016.

\bibitem[Hern{\'a}ndez-Hern{\'a}ndez et~al.(2016)Hern{\'a}ndez-Hern{\'a}ndez, Garc{\'\i}a-Mateos, Gonz{\'a}lez-Esquiva, Escarabajal-Henarejos, Ruiz-Canales, and Molina-Mart{\'\i}nez]{hernandez2016optimal}
Jos{\'e}~Luis Hern{\'a}ndez-Hern{\'a}ndez, Gin{\'e}s Garc{\'\i}a-Mateos, JM Gonz{\'a}lez-Esquiva, D Escarabajal-Henarejos, Antonio Ruiz-Canales, and Jos{\'e}~Miguel Molina-Mart{\'\i}nez.
\newblock Optimal color space selection method for plant/soil segmentation in agriculture.
\newblock \emph{Computers and Electronics in Agriculture}, 122:\penalty0 124--132, 2016.

\bibitem[Jintasuttisak et~al.(2025)Jintasuttisak, Chabplan, Issaro, Saeung, and Suwanroj]{jintasuttisak2025accurate}
Thani Jintasuttisak, Patompong Chabplan, Sasitorn Issaro, Orawan Saeung, and Thamasan Suwanroj.
\newblock Accurate segmentation of vegetation in uav desert imagery using hsv-glcm features and svm classification.
\newblock \emph{Journal of Imaging}, 12\penalty0 (1):\penalty0 9, 2025.

\bibitem[Jocher et~al.(2023)Jocher, Chaurasia, and Qiu]{jocher2023ultralytics}
Glenn Jocher, Ayush Chaurasia, and Jing Qiu.
\newblock Ultralytics yolo.
\newblock \emph{Zenodo}, 2023.

\bibitem[Ling and Okada(2007)]{ling2007efficient}
Haibin Ling and Kazunori Okada.
\newblock An efficient earth mover's distance algorithm for robust histogram comparison.
\newblock \emph{IEEE transactions on pattern analysis and machine intelligence}, 29\penalty0 (5):\penalty0 840--853, 2007.

\bibitem[Lipton(2018)]{lipton2018mythos}
Zachary~C Lipton.
\newblock The mythos of model interpretability.
\newblock \emph{Communications of the ACM}, 61\penalty0 (10):\penalty0 36--43, 2018.

\bibitem[Losing et~al.(2018)Losing, Hammer, and Wersing]{losing2018incremental}
Viktor Losing, Barbara Hammer, and Heiko Wersing.
\newblock Incremental on-line learning: A review and comparison of state of the art algorithms.
\newblock \emph{Neurocomputing}, 275:\penalty0 1261--1274, 2018.

\bibitem[Manfreda et~al.(2018)Manfreda, McCabe, Miller, Lucas, Pajuelo~Madrigal, Mallinis, Ben~Dor, Helman, Estes, Ciraolo, et~al.]{manfreda2018use}
Salvatore Manfreda, Matthew~F McCabe, Pauline~E Miller, Richard Lucas, Victor Pajuelo~Madrigal, Giorgos Mallinis, Eyal Ben~Dor, David Helman, Lyndon Estes, Giuseppe Ciraolo, et~al.
\newblock On the use of unmanned aerial systems for environmental monitoring.
\newblock \emph{Remote sensing}, 10\penalty0 (4):\penalty0 641, 2018.

\bibitem[Pettorelli et~al.(2005)Pettorelli, Vik, Mysterud, Gaillard, Tucker, and Stenseth]{pettorelli2005using}
Nathalie Pettorelli, Jon~Olav Vik, Atle Mysterud, Jean-Michel Gaillard, Compton~J Tucker, and Nils~Chr Stenseth.
\newblock Using the satellite-derived ndvi to assess ecological responses to environmental change.
\newblock \emph{Trends in ecology \& evolution}, 20\penalty0 (9):\penalty0 503--510, 2005.

\bibitem[Rouse~Jr et~al.(1974)Rouse~Jr, Haas, Deering, Schell, and Harlan]{rouse1974monitoring}
John~W Rouse~Jr, R~Hect Haas, DW Deering, JA Schell, and James~C Harlan.
\newblock Monitoring the vernal advancement and retrogradation (green wave effect) of natural vegetation.
\newblock Technical report, 1974.

\bibitem[Rubner et~al.(2000)Rubner, Tomasi, and Guibas]{rubner2000earth}
Yossi Rubner, Carlo Tomasi, and Leonidas~J Guibas.
\newblock The earth mover's distance as a metric for image retrieval.
\newblock \emph{International journal of computer vision}, 40\penalty0 (2):\penalty0 99--121, 2000.

\bibitem[Shi and Yang(2015)]{shi2015support}
Dee Shi and Xiaojun Yang.
\newblock Support vector machines for land cover mapping from remote sensor imagery.
\newblock In \emph{Monitoring and Modeling of Global Changes: A Geomatics Perspective}, pages 265--279. Springer, 2015.

\bibitem[Shi et~al.(2025)Shi, Liao, Wang, Ye, Wang, Yue, and Liu]{shi2025evaluation}
Weibo Shi, Xiaohan Liao, Shaoqiang Wang, Huping Ye, Dongliang Wang, Huanyin Yue, and Jianli Liu.
\newblock Evaluation of a cnn model to map vegetation classification in a subalpine coniferous forest using uav imagery.
\newblock \emph{Ecological informatics}, 87:\penalty0 103111, 2025.

\bibitem[Snell et~al.(2017)Snell, Swersky, and Zemel]{snell2017prototypical}
Jake Snell, Kevin Swersky, and Richard Zemel.
\newblock Prototypical networks for few-shot learning.
\newblock \emph{Advances in neural information processing systems}, 30, 2017.

\bibitem[Song et~al.(2023)Song, Lu, Ding, Sun, Jia, and Sun]{song2023new}
Zhenqi Song, Yuefeng Lu, Ziqi Ding, Dengkuo Sun, Yuanxin Jia, and Weiwei Sun.
\newblock A new remote sensing desert vegetation detection index.
\newblock \emph{Remote Sensing}, 15\penalty0 (24):\penalty0 5742, 2023.

\bibitem[Stamatopoulos et~al.(2024)Stamatopoulos, Le, and Daver]{stamatopoulos2024uav}
I Stamatopoulos, TC Le, and Fugen Daver.
\newblock Uav-assisted seeding and monitoring of reforestation sites: a review.
\newblock \emph{Australian Forestry}, 87\penalty0 (2):\penalty0 90--98, 2024.

\bibitem[Sural et~al.(2002)Sural, Qian, and Pramanik]{sural2002segmentation}
Shamik Sural, Gang Qian, and Sakti Pramanik.
\newblock Segmentation and histogram generation using the hsv color space for image retrieval.
\newblock In \emph{Proceedings. international conference on image processing}, pages II--II. IEEE, 2002.

\bibitem[Suthaharan(2016)]{suthaharan2016support}
Shan Suthaharan.
\newblock Support vector machine.
\newblock In \emph{Machine learning models and algorithms for big data classification: thinking with examples for effective learning}, pages 207--235. Springer, 2016.

\bibitem[Swain and Ballard(1991)]{swain1991color}
Michael~J Swain and Dana~H Ballard.
\newblock Color indexing.
\newblock \emph{International journal of computer vision}, 7\penalty0 (1):\penalty0 11--32, 1991.

\bibitem[Tan and Le(2019)]{tan2019efficientnet}
Mingxing Tan and Quoc Le.
\newblock Efficientnet: Rethinking model scaling for convolutional neural networks.
\newblock In \emph{International conference on machine learning}, pages 6105--6114. PmLR, 2019.

\bibitem[Tharwat(2019)]{tharwat2019parameter}
Alaa Tharwat.
\newblock Parameter investigation of support vector machine classifier with kernel functions: A. tharwat.
\newblock \emph{Knowledge and Information Systems}, 61\penalty0 (3):\penalty0 1269--1302, 2019.

\bibitem[Uzhinskiy et~al.(2021)Uzhinskiy, Ososkov, Goncharov, Nechaevskiy, and Smetanin]{uzhinskiy2021one}
Aleksandr~Vladimirovich Uzhinskiy, Gennadii~Alekseevich Ososkov, Pavel~Vladimirovich Goncharov, Andrei~Vasil'evich Nechaevskiy, and Artem~Alekseevich Smetanin.
\newblock One-shot learning with triplet loss for vegetation classification tasks.
\newblock \emph{Computer Optics (In Russian)}, 45\penalty0 (4):\penalty0 608--614, 2021.

\bibitem[Wang et~al.(2026)Wang, Chen, Wei, Yang, Wang, Liu, Cao, Zhao, Chang, Wang, et~al.]{wang2026uav}
Sheng Wang, Kexin Chen, Ziheng Wei, Linzhe Yang, Qiao Wang, Mingzhi Liu, Kangjian Cao, Chaoyue Zhao, Runlong Chang, Zhen Wang, et~al.
\newblock Uav-based deep learning for biodiversity monitoring: Advances, applications, and future directions.
\newblock \emph{Ecological Informatics}, page 103710, 2026.

\bibitem[Warden and Situnayake(2019)]{warden2019tinyml}
Pete Warden and Daniel Situnayake.
\newblock \emph{TinyML}.
\newblock O'Reilly Media Incorporated: Sebastopol, CA, USA, 2019.

\bibitem[Wenbo et~al.(2025)Wenbo, Peng, Honglei, Wei, and Fuqiang]{wenbo2025identification}
Yang Wenbo, Guan Peng, Shi Honglei, Zhang Wei, and Lei Fuqiang.
\newblock Identification of grassland vegetation coverage and height based on vegetation index and hsv space.
\newblock \emph{Journal of Resources and Ecology}, 16\penalty0 (4):\penalty0 933--945, 2025.

\bibitem[Yang et~al.(2015)Yang, Wang, Zhao, Zhang, and Feng]{yang2015greenness}
Wenzhu Yang, Sile Wang, Xiaolan Zhao, Jingsi Zhang, and Jiaqi Feng.
\newblock Greenness identification based on hsv decision tree.
\newblock \emph{Information Processing in Agriculture}, 2\penalty0 (3-4):\penalty0 149--160, 2015.

\bibitem[Zhang et~al.(2024)Zhang, Bi, and Xuan]{zhang2024convolutional}
Tao Zhang, Yuge Bi, and Chuanzhong Xuan.
\newblock Convolutional transformer attention network with few-shot learning for grassland degradation monitoring using uav hyperspectral imagery.
\newblock \emph{International Journal of Remote Sensing}, 45\penalty0 (6):\penalty0 2109--2135, 2024.

\end{thebibliography}
}

\end{document}